\documentclass{article}

\usepackage{microtype}
\usepackage{graphicx}
\usepackage{subfigure}
\usepackage{booktabs} 
\usepackage{amsfonts}

\usepackage{hyperref}

\usepackage[accepted]{sysml2019}

\sysmltitlerunning{Lightweight Convolutional Representations for On-Device Natural Language Processing}

\begin{document}

\twocolumn[
\sysmltitle{Lightweight Convolutional Representations\\for On-Device Natural Language Processing}



\begin{sysmlauthorlist}
\sysmlauthor{Shrey Desai}{ut}
\sysmlauthor{Geoffrey Goh}{fb}
\sysmlauthor{Arun Babu}{fb}
\sysmlauthor{Ahmed Aly}{fb}
\end{sysmlauthorlist}

\sysmlaffiliation{ut}{The University of Texas at Austin}
\sysmlaffiliation{fb}{Facebook Assistant}

\sysmlcorrespondingauthor{Shrey Desai}{shreydesai@utexas.edu}
\sysmlcorrespondingauthor{Geoffrey Goh}{geoffreygoh@fb.com}
\sysmlcorrespondingauthor{Arun Babu}{arbabu@fb.com}
\sysmlcorrespondingauthor{Ahmed Aly}{ahhegazy@fb.com}

\sysmlkeywords{Computation and Language, Machine Learning, SysML}

\vskip 0.3in

\begin{abstract}
The increasing computational and memory complexities of deep neural networks have made it difficult to deploy them on low-resource electronic devices (e.g., mobile phones, tablets, wearables). Practitioners have developed numerous model compression methods to address these concerns, but few have condensed input representations themselves. In this work, we propose a fast, accurate, and lightweight convolutional representation that can be swapped into any neural model and compressed significantly (up to 32x) with a negligible reduction in performance. In addition, we show gains over recurrent representations when considering resource-centric metrics (e.g., model file size, latency, memory usage) on a Samsung Galaxy S9.
\end{abstract}
]



\printAffiliationsAndNotice{} 

\section{Introduction}

Recent developments in deep learning for natural language processing (NLP) have increased requirements on compute and memory, making it difficult to deploy existing models on low-resource electronic devices due to sharp hardware constraints. This poses a need for developing a new class of models that use a fraction of these resources during inference while matching existing performance.

In recent years, convolutional neural networks (CNNs) have become ubiquitous for NLP tasks \cite{collobert2011natural, kim2014convolutional}. Stacked convolutions build complex representations from low-level signals, making them useful for language tasks. Previous work employing convolutional architectures have shown significant gains over recurrent counterparts both in accuracy and speed \cite{gehring2017convolutional, dauphin2017gated}. CNNs exhibit high degrees of parallelism against the need for sequential unrolling in RNNs. Dedicated digital signal processors (DSPs), field programmable gate array (FPGA) accelerators, and mobile GPUs have also been designed to speed up CNNs on-device \cite{kangaccelerator2019, solovyev2018fpga, hunynh2017}. Combined with co-optimization techniques like neural architecture search, quantization, and pruning, CNN models can run significantly faster \cite{zoph2017neural, lin2016fixed, li2017pruning}.

In this paper, we develop a fast, accurate, and compact convolutional representation that can be plugged into any model. Our representation is organized in residual blocks \cite{he2015deep}, where each block is composed of a series of lightweight convolutions followed by a non-linearity. In contrast to architecture-specific model compression methods, it can be incorporated into any neural encoder that forms representations of the input, whether it be at the word-, character-, or byte-level. We comprehensively evaluate its performance by comparing it against recurrent representations on three on-device NLP tasks: next word prediction, joint intent-slot modeling, and document classification. Lastly, we benchmark their runtime performance on a Samsung Galaxy S9 using resource-centric metrics like model file size, latency and memory consumption.

\section{Representation}

Our convolutional representation builds on top of the convolutional encoder described in \citet{gehring2017convolutional}. This encoder ingests input embeddings, then processes them through $n$ blocks. Each block consists of a dropout layer, convolutional layer, non-linearity, as well as a residual connection \cite{he2015deep} from its input to its output to enable training a deeper network. We also introduce and detail several techniques which make this representation lighter in the following sections.

\subsection{Non-Linearity} \label{repr:nl}

\citet{gehring2017convolutional} use Gated Linear Unit (GLU) activations in their encoder. GLUs mimic the gating behavior of recurrent networks by performing two groups of convolutions over input embeddings and then combining the output of the first with the sigmoid of the second. Although convolutional gates are performant, GLUs double the number of required output channels, adding unnecessary capacity to the model. Through experimentation, we evaluated several cheaper non-linearities, such as ReLU \cite{nair2010rectified}, Leaky ReLU \cite{xu2015empirical}, Exponential Linear Units (ELU) \cite{clevert2016fast}, and Gaussian Error Linear Units (GELU) \cite{hendrycks2018gaussian}, and found GELU to work best in terms of accuracy and speed.

\subsection{Depthwise Separable Convolutions} \label{repr:separability}

The standard convolution operation requires a significant number of operations to compute, and serially stacking multiple blocks exacerbates the problem. To illustrate, let $c_i$ denote the number of input channels to the $i$th (one-dimensional) convolution and $t_i$ denote the length of the feature map. The $i$th convolution consumes an input feature map $\mathbf{x}_i \in \mathbb{R}^{c_i \times t_i}$. It uses $c_{i+1}$ filters to transform this into an output feature map $\mathbf{x}_{i+1} \in \mathbb{R}^{c_{i+1} \times t_{i+1}}$, where each filter $f_{ij} \in \mathbb{R}^{c_i \times k}$ consists of $c_i$ kernels of size $k$. For simplicity, we keep the number of input and output channels the same, thus each convolution requires $c^2kt_{i+1}$ operations to compute. Since $k$ is relatively small and the dimensionality of $t$ is successively reduced, the number of operations is dominated by the channel dimension $c$. 

Inspired by the Xception architecture for image classification \cite{chollet2017xception}, we use depthwise separable convolutions to reduce the number of operations in computing a convolution. Traditional convolutions jointly learn to extract spatial features and transition them into a new channel space. However, this process can be decoupled into two smaller, simpler steps: (1) the \textit{depthwise} convolution spatially convolves the input; (2) the \textit{pointwise} convolution (usually with a $1 \times 1$ kernel) projects the result of the depthwise convolution into a new channel space. When performed together, this results in a total of $c^2kt_{i+1} + c^2t_{i+1}$ operations, corresponding to the number of operations required to compute a depthwise and pointwise convolution, respectively. However, without further optimization, the number of total operations \textit{increases} over those in the traditional convolution by $c^2t_{i+1}$. We address this issue in two ways.

\paragraph{Separability} First, we optimize the complexity of the depthwise convolution by leveraging grouped convolutions \cite{krizhevsky2012imagenet}, which split the $c$ input channels into $g$ disjoint groups, spatially convolve each group separately, and concatenate the results. As a result, each depthwise filter $f_{ij}^d \in \mathbb{R}^{\frac{c}{g} \times k}$ contains a factor of $g$ less kernels. \citet{krizhevsky2012imagenet} use $g=2$ to learn two distinct filter groups. In contrast, we use extreme grouping by setting $g=c$, defining a depthwise filter $f_{j}^d \in \mathbb{R}^k$ that convolves each input channel with \textit{one} kernel. This reduces the number of depthwise convolution operations by a factor of $c$, yielding a total operation count of $ckt_{i+1} + c^2t_{i+1}$.

\paragraph{Bottlenecks} Second, we optimize the complexity of the pointwise convolution, which requires a quadratic number of operations with respect to $c$. We introduce a bottleneck dimension $b$, and further decompose the large pointwise convolution into two smaller sub-convolutions: (1) the \textit{downsampling} convolution projects the $c$-dimensional depthwise convolutional representation into a $b$-dimensional space; (2) the \textit{upsampling} convolution projects the obtained $b$-dimensional representation into the target, $c$-dimensional space. Therefore, we split the pointwise filter into two filters $f_{ij}^{p_d} \in \mathbb{R}^{c \times b}$ and $f_{ij}^{p_u} \in \mathbb{R}^{b \times c}$ where $p_d$ and $p_u$ represent the downsampling and upsampling pointwise convolutions, respectively. If $b$ is parameterized such that $2b < c$, the total number of parameters across filters $f_{ij}^{p_d}$ and $f_{ij}^{p_u}$ will be diminished. Under this formulation, the total number of operations computed by an optimized convolution becomes $ckt_{i+1} + 2bct_{i+1}$, a large reduction over the original count.

\section{Experiments} \label{tasks:intro}

We perform an extensive evaluation of our convolutional representation in comparison to recurrent representations on three on-device NLP tasks: next word prediction (\S\ref{task:nwp}), joint intent slot modeling (\S\ref{task:jis}), and document classification (\S\ref{task:dc}). We also perform ablations to demonstrate the relative importance of the non-linearity, separability, and bottleneck components in our representation. 

\subsection{Next Word Prediction} \label{task:nwp}

The next word prediction task involves predicting the continuation of a sequence (i.e., the next word) given the sequence itself. For example, a keyboard with such capabilities can give recommendations for the next word to type as a message is being written. This task largely collapses to learning a language model, where during inference, we select the word $w$ that maximizes the conditional probability $p(w|x_0, \cdots, x_{i-1})$ of a sequence.

\paragraph{Dataset} We use a dataset consisting of sentences extracted from various social media websites \cite{yu2018ondevice}. These sentences are largely informal, so they mimic the type of phrases that would typed on a keyboard. Consistent with the original paper, we use 5.8M/968.2K/2.9M sentences for train/dev/test and 15K vocabulary words. In addition to perplexity, we use keystroke savings (percentage of keystrokes a user does not press due to typeahead) and word prediction rate (how many words a model predicts correctly in a test set) to evaluate the language model, both using the manually curated test dataset provided by \citet{yu2018ondevice}.

\paragraph{Model} Our model architecture consists of an embedding layer $\mathrm{E}$, representation layer $\mathrm{R}$ (recurrent or convolutional), and linear layer $\mathrm{L}$. To reduce the number of free parameters, we decompose $\mathrm{E}$'s weight matrix $\mathbf{W}_e$ into the product of two sub-matrices $\mathbf{W}_e^a$ and $\mathbf{W}_e^b$. Similarly, we decompose $\mathrm{L}$'s weight matrix $\mathbf{W}_\ell$ as $\mathbf{W}_\ell^a$ and $\mathbf{W}_\ell^b$. We tie $\mathbf{W}_e^a$ with $\mathbf{W}_\ell^a$ and $\mathbf{W}_e^b$ with $\mathbf{W}_\ell^b$. This decomposition is primarily motivated by the reduction in file size when storing two significantly smaller sub-matrices. Each forward pass consists of the following steps: (1) we create the embedding matrix ($\mathbf{W}_e \leftarrow \mathbf{W}_e^a \times \mathbf{W}_e^b$) and use it to ingest $d$-dimensional embeddings; (2) $\mathrm{R}$ creates a representation  of the sequence; (3) we create the linear matrix ($\mathbf{W}_\ell \leftarrow \mathbf{W}_\ell^a \times \mathbf{W}_\ell^b$) and use it to obtain a (non-normalized) distribution over the vocabulary.

\begin{table}[t]
\centering
\small
\begin{tabular}{l|c|c|c|c}
\toprule
Representation & Parameters & PPL & KS & WPR \\ \midrule
Recurrent & 2.2 M & 57.2 & 66.3 & 39.1 \\
Convolutional & 3.6 M & \textbf{56.2} & \textbf{66.7} & \textbf{39.7} \\
~~+ non-linearity & 2.8 M & 58.1 & 66.0 & 38.6 \\
~~~~+ separability & 2.2 M & 59.2 & 66.1 & 38.9 \\
~~~~~~+ bottlenecks & \textbf{2.1 M} & 61.2 & 65.9 & 38.6 \\
\bottomrule
\end{tabular}
\caption{Next word prediction results, reporting number of parameters, word-level perplexity (PPL), keystroke savings (KS), and word prediction rate (WPR).}
\label{res:nwp-res}
\end{table}

\begin{table}[t]
\centering
\small
\begin{tabular}{l|c|c|c}
\toprule
Representation & File Size & Latency & Memory \\ \midrule
Recurrent & 11.4 MB & 33.0 ms & 126.5 MB \\
Convolutional & 18.2 MB & 31.1 ms & 141.9 MB \\
~~+ non-linearity & 14.2 MB & 26.1 ms & 132.0 MB \\
~~~~+ separability & 11.5 MB & 22.7 ms & 124.7 MB \\
~~~~~~+ bottlenecks & \textbf{11.0 MB} & \textbf{22.2 ms} & \textbf{123.5 MB} \\
\bottomrule
\end{tabular}
\caption{Next word prediction benchmarks, reporting model file size, latency, and memory usage on a Samsung Galaxy S9.}
\label{res:nwp-bench}
\end{table}

\paragraph{Results} Table \ref{res:nwp-res} displays our model results. Our optimized representation is smaller than the recurrent and convolutional baselines with a minor reduction in performance. We found word-level perplexity to be a poor measure of performance for this task, as the 8\% increase in perplexity is mostly due to the language model's inability to shape accurate distributions for rare words (e.g., r/bitcoin, rondo, dpi)---a byproduct of the reduction in capacity. These words do not usually appear in day-to-day text messages, explaining why the keystroke savings (KS) and word prediction rate (WPR) metrics do not suffer much. Table \ref{res:nwp-bench} displays our benchmark results. Our optimized representation shows improvements over the baseline recurrent and convolutional representations across all on-device metrics. In particular, it brings down the latency by 32\% in comparison to the recurrent model.

\subsection{Joint Intent Slot Modeling} \label{task:jis}

Intent classification and slot filling create semantic parses for spoken language utterances. Intent classification is a document classification task that assigns a categorical label $y^{\ell}_i$ given utterance tokens $(x_1, \cdots, x_T)$. Slot filling is a sequence labeling task that tags each utterance token $(x_1, \cdots, x_T)$ with the associated slots $(y^s_1, \cdots, y^s_T)$. Both tasks are commonly jointly learned \cite{goo2018}.

\paragraph{Dataset} We use a manually curated dataset of device control utterances common across smart devices, spanning 36 intent classes and 14 slot tags. The utterances are hand-designed and verified by a team of linguists; no user data or information was used in the creation of this dataset. They are designed either to open apps (e.g., ``please open spotify") or tune device settings (e.g., ``turn up the volume"). Each is also annotated with a number of gazetteer features, specifically tags for named entities. We use 36.2K/4.24K/10.6K samples for train/dev/test.

\paragraph{Model} Our model architecture consists of a character embedding layer $\mathrm{E}_c$, gazzetteer embedding layer $\mathrm{E}_g$, intent representation layer $\mathrm{R}_i$, slot representation layer $\mathrm{R}_s$, intent linear layer $\mathrm{L}_i$, and slot linear layer $\mathrm{L}_s$. $\mathrm{R}_i$ and $\mathrm{R}_s$ are parameterized as either recurrent or convolutional representations. $\mathrm{E}_c$ learns character embeddings, then uses multiple convolutions and max pooling to form a representation for each word \cite{kim2015character}. $\mathrm{E}_g$ learns embeddings for each gazetteer features, then uses max pooling to form a gazetteer representation for each word. Each forward pass consists of the following steps: (1) $\mathrm{E}_c$ and $\mathrm{E}_g$ obtain word-level representations and are concatenated; (2) $\mathrm{R}_i$ and $\mathrm{R}_s$ create intent and slot representations, respectively; (3) $\mathrm{L}_i$ and $\mathrm{L}_s$ decode the intent and slot representations, respectively, into the class space.

\begin{table}
\centering
\small
\begin{tabular}{l|c|c|c}
\toprule
Representation & Parameters & Intent F1 & Slot F1 \\ \midrule
Recurrent & 1.4 M & 95.0 & \textbf{94.8} \\
Convolutional & 1.1 M & 96.1 & 93.6 \\
~~+ non-linearity & 719.5 K & 95.3 & 94.1 \\
~~~~+ separability & 417.2 K & \textbf{96.2} & 94.3 \\
~~~~~~+ bottlenecks & \textbf{292.3 K} & 95.3 & 94.6 \\
\bottomrule
\end{tabular}
\caption{Joint intent slot modeling results, reporting number of parameters and micro F1 scores for intent slot classification and slot filling.}
\label{res:jis-res}
\end{table}

\begin{table}
\centering
\small
\begin{tabular}{l|c|c|c}
\toprule
Representation & File Size & Latency & Memory \\ \midrule
Recurrent & 7.0 MB & 27.7 ms & 13.1 MB \\
Convolutional & 5.7 MB & 6.6 ms & 14.1 MB \\
~~+ non-linearity & 3.5 MB & 5.0 ms & 9.3 MB \\
~~~~+ separability & 2.0 MB & 4.1 ms & 6.2 MB \\
~~~~~~+ bottlenecks & \textbf{1.4 MB} & \textbf{3.7 ms} & \textbf{5.3 MB} \\
\bottomrule
\end{tabular}
\caption{Joint intent slot modeling benchmarks, reporting model file size, latency, and memory usage on a Samsung Galaxy S9.}
\label{res:jis-bench}
\end{table}

\paragraph{Results} Table \ref{res:jis-res} displays our model results. Our optimized representation shows a modest amount of compression, approximately 4-4.8x smaller than the baselines in the number of parameters alone. The intent classification and slot filling micro F1 scores also stay relatively constant---both drop by less than 1\%. Table \ref{res:jis-bench} displays our benchmark results. In comparison to the recurrent baseline, we see a 80\% reduction in file size, 86\% reduction in latency, and 59\% reduction in memory.

\subsection{Document Classification} \label{task:dc}

The document classification task assigns each document $x_i$ a categorical label $y_i$. We represent documents at the byte-level to reduce the footprint of our on-device model. For simplification, we remove sentence boundaries and treat the document as a continuous sequence of words. Each word $w_i$ is encoded as a sequence of bytes, where each byte $b_{ij}$ is one of 256 extended ASCII character encodings.

\paragraph{Dataset} We use the Books, Electronics, Movies, CDs, and Home categories of the Amazon Product Reviews dataset \cite{he2016}. The dataset contains fine-grained sentiment labels, ranking product reviews from 1-5. We create positive (4/5) and negative (1/2) categories, discarding the neutral (3) reviews. We use 100K/50K/50K documents for train/dev/test.

\paragraph{Model} Our model architectures consists of an embedding layer $\mathrm{E}$, representation layer $\mathrm{R}$ (recurrent or convolutional), and linear layer $\mathrm{L}$. Each forward pass consists of the following steps: (1) $\mathrm{E}$ returns $d$-dimensional byte-level embeddings; (2) $\mathrm{R}$ creates a representation of the input, which is subsequently max or average pooled; (3) $\mathrm{L}$ projects the sequence representation into the binary class space.

\begin{table}[t]
\centering
\small
\begin{tabular}{l|c|c}
\toprule
Representation & Parameters & Accuracy \\ \midrule
Recurrent & 509.6 K & \textbf{87.4} \\
Convolutional & 2.5 M & 85.7 \\
~~+ non-linearity & 1.3 M & 87.3 \\
~~~~+ separability & 234.4 K & 86.7 \\
~~~~~~+ bottlenecks & \textbf{92.5 K} & 86.4 \\
\bottomrule
\end{tabular}
\caption{Document classification results, reporting number of parameters and accuracy.}
\label{dc:dc-res}
\end{table}

\begin{table}[t]
\centering
\small
\begin{tabular}{l|c|c|c}
\toprule
Representation & File Size & Latency & Memory \\ \midrule
Recurrent & 2.8 MB & 109.9 ms & \textbf{6.5 MB} \\
Convolutional & 12.8 MB & 92.4 ms & 40.3 MB \\
~~+ non-linearity & 6.5 MB & 50.3 ms & 22.7 MB \\
~~~~+ separability & 1.1 MB & 15.0 ms & 10.4 MB \\
~~~~~~+ bottlenecks & \textbf{0.4 MB} & \textbf{10.1 ms} & 8.9 MB \\
\bottomrule
\end{tabular}
\caption{Document classification benchmarks, reporting model file size, latency, and memory usage on a Samsung Galaxy S9.}
\label{dc:dc-bench}
\end{table}

\paragraph{Results} Table \ref{dc:dc-res} displays our model results. Our optimized representation is significantly smaller than the baselines, using 5x and 27x less parameters than the recurrent and convolutional baselines respectively. Even with far less capacity, the optimized model's accuracy is comparable to the baselines. Specifically, it only drops by 1\% in comparison to the recurrent baseline. In addition, each successively compressed convolutional model achieves an accuracy better than the baseline. This suggests that the original convolutional model is over-parameterized and the reduction in capacity helps during optimization. Table \ref{dc:dc-res} displays our benchmark results. Our optimized model shows gains on the file size and latency metrics; most notably, its latency is an order of magnitude lower than the recurrent baseline's. However, the CNNs all use slightly more memory than the recurrent model, presenting a clear trade-off between latency and memory requirements.

\section{Related Work}

Practitioners have developed general methods for compressing neural networks, including knowledge distillation \cite{hinton2015distilling}, quantization \cite{gong2014compressing, hubara2017quantized}, and pruning \cite{zhu2017prune, liu2018rethinking, gale2019state}. Our representation can be used in conjunction with these methods. For example, the model in our language modeling task uses matrix decomposition alongside our lightweight representation. Architecture-specific compression techniques have also been proposed. \citet{wen2018learning} learn intrinsic sparse structures in LSTMs; \citet{li2017pruning} use the $\ell_1$ norm to structurally prune convolutional filters; and \citet{michel2019are} iteratively prune attention heads in self-attention models. In contrast, our representation is not unique to any one model architecture; our experiments point towards generalizability across multiple inputs, models, and tasks. Moreover, we demonstrate the applicability of our techniques in a real-world scenario.

\section{Conclusion}

In this paper, we introduce a new lightweight convolutional representation for on-device NLP that outperform strong baselines (e.g., recurrent representations) in terms of model file size, latency, and memory consumption. It (1) works with word-, character-, and byte-level inputs; (2) can be switched into any neural model that forms representations of the input; and (3) is generalizable across three NLP tasks: next word prediction, joint intent slot modeling, and document classification. Additionally, the runtime performance of our fully optimized models prove their applicability for commercial use cases. Future work will explore the viability of this representation in more language tasks.


\bibliography{example_paper}
\bibliographystyle{sysml2019}

%


\end{document}